\documentclass[sigconf]{acmart}
\AtBeginDocument{%
  }

\setcopyright{acmlicensed}
\copyrightyear{2026}
\acmYear{2026}
\acmDOI{XXXXXXX.XXXXXXX}
\acmConference[]{International Conference on Artificial Intelligence and Law}{June 2026}{Woodstock, NY}
\acmISBN{978-1-4503-XXXX-X/2018/06}

\begin{document}

\title[Tree-of-Thoughts Inspired Legal Summarization using LLMs]{A Tree-of-Thoughts Inspired Hybrid Approach for Legal Case Judgement Summarization using LLMs}


\author{Aniket Deroy}
\email{roydanik18@gmail.com}
\affiliation{%
  \institution{IIT Kharagpur}
  \country{India}
}

\author{Kripabandhu Ghosh}
\email{kripaghosh@iiserkol.ac.in}
\affiliation{%
  \institution{IISER Kolkata}
  \country{India}
}

\author{Saptarshi Ghosh}
\email{saptarshi@cse.iitkgp.ac.in}
\affiliation{%
  \institution{IIT Kharagpur}
  \country{India}
}




\begin{abstract}
In recent times, Large Language Models
(LLMs) are increasingly being used for legal case judgement summarization. 
Most prior works have tried traditional extractive and abstractive summarization of case judgements. However, hybrid or  extractive-abstractive techniques have not been explored much.
In this work, we propose a novel \textit{tree-of-thoughts} inspired extractive-abstractive summarization approach for legal judgement summarization.
We conduct experiments using two popular LLMs, DeepSeek and LLama, and compare among extractive, abstractive and extractive-abstractive summarization. 
Our experiments show that the proposed extractive-abstractive prompt
provides better summaries compared to other types of LLM prompts. 
\end{abstract}

\maketitle

%

\section{Introduction}

Summarizing legal judgments is a crucial and complex task that has been attempted for several decades. Legal judgments are typically lengthy, intricate, and filled with domain-specific terminology and reasoning. Automated summarization systems aim to condense these documents while preserving essential information such as legal arguments, factual details, and judgments.
Traditionally, two types of summaries have been mostly explored in literature.
\textit{Extractive summarization} acts like a highlighter, copying original sentences to ensure factual accuracy but often resulting in an unnatural flow of the summary text. 
Whereas, \textit{abstractive summarization} acts like a writer, paraphrasing ideas into new, natural-sounding sentences, though it risks creating factual errors.

Recent advancements in Large Language Models 
(LLMs) have shown remarkable performance in generating both extractive and abstractive summaries of legal documents.
However, to our knowledge, there has not been much systematic exploration of hybrid (extractive-abstractive) approaches for summarizing legal judgments using LLMs. 

In this work, we \textit{propose a tree-of-thoughts inspired extractive-abstractive approach} for case judgement summarization using LLMs, (which we call \textbf{Ext-Abs-ToT}). This approach aims to combine the strengths of extractive and abstractive approaches, by first extracting important parts from the document, and then generating a coherent summary. 

We compare the three summarization approaches -- extractive, abstractive, and extractive-abstractive -- using two popular LLMs (DeepSeek and Llama) over a standard legal case summarization dataset. 
We evaluate the quality of the generated summaries using several standard metrics such as Rouge-2, Rouge-L, METEOR,  MoverScore and BERTScore. 
Our experiments show that the proposed extractive-abstractive approach generates better quality summaries than the traditional extractive and abstractive approaches. 
This study shows that hybrid extractive-abstractive approaches are promising and should be explored further in the future.


\begin{table}[tb]
\centering
\begin{tabular}{|p{1.5cm}|p{6.4cm}|}
\hline
Type & Prompt used\\
\hline
Extractive & Generate an extractive summary from the legal document - $<$~legal document~$>$. Choose the most important sentences from the legal document to form the legal document summary. \\ \hline

Abstractive & Generate an abstractive summary from the legal document - $<$~legal document~$>$. Generate the information in the summary in a readable, coherent and natural manner while preserving the core meaning and important details. \\ \hline

\hline
\end{tabular}
\caption{Baseline prompts for extractive and abstractive summarization}
\label{tab:prompts}
\end{table}

\section{Prompts for summarization}

We use two baseline prompts for extractive and abstractive summarization via LLMs, as shown in Table~\ref{tab:prompts}.

We propose a tree-of-thoughts inspired prompting approach for extractive-abstractive summarization, which we refer to as the extractive-abstractive-ToT (\textbf{Ext-Abs-ToT}) approach.
The process is a 3-stage prompting approach, as shown in
Table~\ref{tab:prompts_tot}. 
Stage~1 extracts important parts from the input document, stage~2 estimates the importance of the extracted parts, and stage~3 generates a summary from those important parts.

\section{Experimental Setup}

\noindent \textbf{Datasets:}
For our experiments, we use the \textbf{IN-Ext dataset} for legal case judgement summarization developed by the prior work~\cite{shukla-etal-2022-legal}. 
The datasets consists of 50 (legal judgement, gold standard summary) pairs where the summaries are created by Law experts from a reputed Law school in India. 

\vspace{2mm}
\noindent \textbf{LLMs:}
We use the popular LLMs DeepSeek-R1~\cite{guo2025deepseek} and Llama-3.2~\cite{dubey2024llama} for our experiments. We choose these LLMs since they are long-context LLMs, and hence most legal case judgements can be accommodated at once in their input context. 

\vspace{2mm}
\noindent \textbf{Quality metrics:}
We measure the ROUGE scores~\cite{lin2004rouge}, METEOR~\cite{banerjee2005meteor}, Moverscore~\cite{zhao2019moverscore} and BERTScore~\cite{zhangbertscore} between every model-generated summary and the gold standard summary. 
In particular, for ROUGE scores, we measure the Rouge-2 F1 score and the Rouge-L F1 score.
We use the implementations of the metrics from the following library -\url{https://github.com/Yale-LILY/SummEval}.
All these metrics compare a model-generated summary with a gold standard summary.
We average the scores for all the 50 data points (judgment-summary pairs) in the dataset and report these scores.

\begin{table}[tb]
\centering
\small
\begin{tabular}{|p{0.95\columnwidth}|}
\hline

\textbf{Stage 1 Prompt:} You are an expert legal summarizer. Your task is to process the following full legal judgment and decompose it into its core rhetorical segments: Facts, Issues Presented, Holding/Ruling, and Court's Reasoning/Ratio Decidendi. For each segment, you must only extract the most salient, supporting sentences from the original text that capture the essence of that segment. Do not generate any new text.

[FULL TEXT OF LEGAL DOCUMENT]

\vspace{2mm}
\textbf{Stage 2 prompt:}
Evaluate the set of extracted snippets provided in the previous step. Assign a confidence score (1-5, 5 being highest) for each of the four segments (Facts, Issues, Holding, Reasoning) based on two criteria: Completeness (Does it capture all necessary points?) and Fidelity (Are all snippets direct, unedited extractions from the original text?). 

[Output from Stage 1]

\vspace{2mm}
\textbf{Stage 3 prompt:}
Using only the validated extractive snippets from the final output of the previous step that is given below, generate a single, coherent, and highly-condensed abstractive summary of the legal judgment. The summary must be written in a formal legal style and seamlessly integrate the Facts, Issue, Holding, and Reasoning into a fluid narrative. Focus on creating novel, non-extractive sentences that preserve the factual and legal integrity of the source material. 
[Output from Stage 2] \\
\hline
\end{tabular}
\caption{Proposed prompt for extractive-abstractive Tree-of-Thoughts summarization}
\label{tab:prompts_tot}
\end{table}

\section{Results}

\begin{table}[t]
\centering
\small
\begin{tabular}{l|ccccc}
\toprule
\textbf{Model}                    &  \textbf{R2-F1}  & \textbf{RL-F1} & \textbf{ME} & \textbf{BS} & \textbf{MS}  \\ 


\midrule
\multicolumn{6}{l}{\textbf{DeepSeek-R1}} \\ 
\midrule

DeepSeek-Ext  &    0.295  & 0.261  & 0.260 & 0.564 & 0.531 \\

DeepSeek-Abs   & 0.249            & 0.262          & 0.240           & \textbf{0.598} & 0.567 \\ 

DeepSeek-Ext-Abs-ToT         & \textbf{0.306}   & \textbf{0.285}          & \textbf{0.275}           & 0.596 & \textbf{0.572}          \\

\midrule
\multicolumn{6}{l}{\textbf{Llama-3.2}} \\ 
\midrule

Llama-Ext    & 0.224  & 0.224  & 0.205           & 0.520 & 0.516      \\ 

Llama-Abs                                            &  0.230       & 0.250          & 0.232 &  0.586 & 0.540 \\

Llama-Ext-Abs-ToT                                                   & \textbf{0.239}                     & \textbf{0.259}          & \textbf{0.239}           & \textbf{0.587} & \textbf{0.548}    \\

\bottomrule

\end{tabular}
\caption{Standard summarization metrics for two LLMs on the IN-Ext dataset. 
We run every LLM with three different prompts: extractive (Ext), abstractive (Abs), and the proposed Ext-Abs-ToT. 
The best value for every metric and a particular LLM is highlighted in boldface.}
\label{tab:automated_metrics_inext}
\end{table}




Table~\ref{tab:automated_metrics_inext} shows the evaluation metrics for the summaries generated by the two LLMs (DeepSeek and Llama). 
Each LLM is applied with three different prompts: extractive (Ext), abstractive (Abs), and the proposed Ext-Abs-ToT.

Across most metrics, the proposed Tree-of-Thoughts approach (Ext-Abs-ToT) performs better than the purely extractive and purely abstractive approaches. 

In general, for the IN-Ext dataset, the DeepSeek-R1 variants perform better than the LLama variants.
Also, the abstractive versions usually get slightly higher scores than the extractive versions.
In turn, the extractive-abstractive versions achieve slightly higher scores than the abstractive versions.

\section{Conclusion}

In this work, we compare three prompting approaches for using LLMs for summarizing legal judgements -- extractive, abstractive, and extractive-abstractive. 
The proposed Tree-of-Thoughts approach shows the highest performance in terms of several standard summarization quality metrics, thus showing that  
the extractive-abstractive summarization approach can capture the good points of both the traditional approaches, and hence is a promising approach to study further.

\vspace{3mm}
\noindent \textbf{Acknowledgments:} This research is
partially supported by the IIT Kharagpur Technology Innovation Hub on AI for Interdisciplinary Cyber-Physical Systems (AI4ICPS) through a project titled ``NyayKosh: Multilingual Resources for AI- based Legal Analytics''.

\bibliographystyle{ACM-Reference-Format}
\bibliography{sample-base}

\end{document}